\documentclass[10pt,twocolumn,letterpaper]{article}

\usepackage{iccv}              

\definecolor{iccvblue}{rgb}{0.21,0.49,0.74}
\usepackage[pagebackref,breaklinks,colorlinks,allcolors=iccvblue]{hyperref}
\usepackage{amsmath}
\usepackage{algorithm}
\usepackage{algorithmic}
\usepackage[accsupp]{axessibility}  

\usepackage{times}
\usepackage{soul}
\usepackage{url}
\usepackage[utf8]{inputenc}
\usepackage{graphicx}
\usepackage{amsmath}
\usepackage{amsthm}
\usepackage{booktabs}
\usepackage[switch]{lineno}
\usepackage{multirow} 
\usepackage{soul}
\usepackage{url}
\usepackage{anyfontsize}
\usepackage{booktabs}
\usepackage[switch]{lineno}
\usepackage{amsfonts}
\usepackage{amssymb}
\usepackage{indentfirst}
\usepackage{bbding}
\usepackage{colortbl}
\usepackage{amsmath}
\usepackage{colortbl}
\usepackage{multirow} 
\usepackage{bbding}
\usepackage{bm}
\definecolor{VibrantPink}{RGB}{255, 20, 147}
\def\paperID{8069} 
\def\confName{ICCV}
\def\confYear{2025}

\makeatother
\title{Few-Shot Image Quality Assessment via Adaptation of Vision-Language Models}

\author{
    Xudong Li$^{1}$\textsuperscript{*} \quad 
    Zihao Huang$^{2}$\textsuperscript{*} \quad 
    Yan Zhang$^{1}$\textsuperscript{\textdagger} \quad  
    Yunhang Shen$^{3}$  \quad 
    Ke Li$^{3}$ \\
    Xiawu Zheng$^{1}$ \quad 
    Liujuan Cao$^{1}$ \quad 
    Rongrong Ji$^{1}$
\\
    $^{1}$ Key Laboratory of Multimedia Trusted Perception and Efficient Computing, \\
    Ministry of Education of China, Xiamen University, 361005, P.R. China\\
$^{2}$ Beijing Institute of Technology \quad
$^{3}$ Tencent Youtu Lab, Shanghai, China \\
\tt\small \{lxd761050753, huangzihhhh, bzhy986\}@gmail.com, 
\tt\small \{zhengxiawu, rrji\}@xmu.edu.cn
}

\begin{document}
\maketitle
{\let\thefootnote\relax\footnote{\textsuperscript{*}Equal contribution. \quad \textsuperscript{†}Corresponding authors.}}

\begin{abstract}
Image Quality Assessment (IQA) remains an unresolved challenge in computer vision due to complex distortions, diverse image content, and limited data availability. Existing Blind IQA (BIQA) methods largely rely on extensive human annotations, which are labor-intensive and costly due to the demanding nature of creating IQA datasets. To reduce this dependency, we propose the Gradient-Regulated Meta-Prompt IQA Framework (GRMP-IQA), designed to efficiently adapt the visual-language pre-trained model, CLIP, to IQA tasks, achieving high accuracy even with limited data. GRMP-IQA consists of two core modules: (i) Meta-Prompt Pre-training Module and (ii) Quality-Aware Gradient Regularization. The Meta Prompt Pre-training Module leverages a meta-learning paradigm to pre-train soft prompts with shared meta-knowledge across different distortions, enabling rapid adaptation to various IQA tasks. On the other hand, the Quality-Aware Gradient Regularization is designed to adjust the update gradients during fine-tuning, focusing the model's attention on quality-relevant features and preventing overfitting to semantic information. 
Extensive experiments on standard BIQA datasets demonstrate the superior performance to the state-of-the-art BIQA methods under limited data setting. Notably, utilizing just {20\%} of the training data, GRMP-IQA is competitive with most existing fully supervised BIQA approaches.
Our code is available via \href{https://github.com/LXDxmu/GRMP-IQA}{\textcolor{iccvblue}{https://github.com/LXDxmu/GRMP-IQA}}.
\end{abstract}

\section{Introduction}
\begin{figure}[t]
  \centering
    \includegraphics[width=0.47\textwidth]{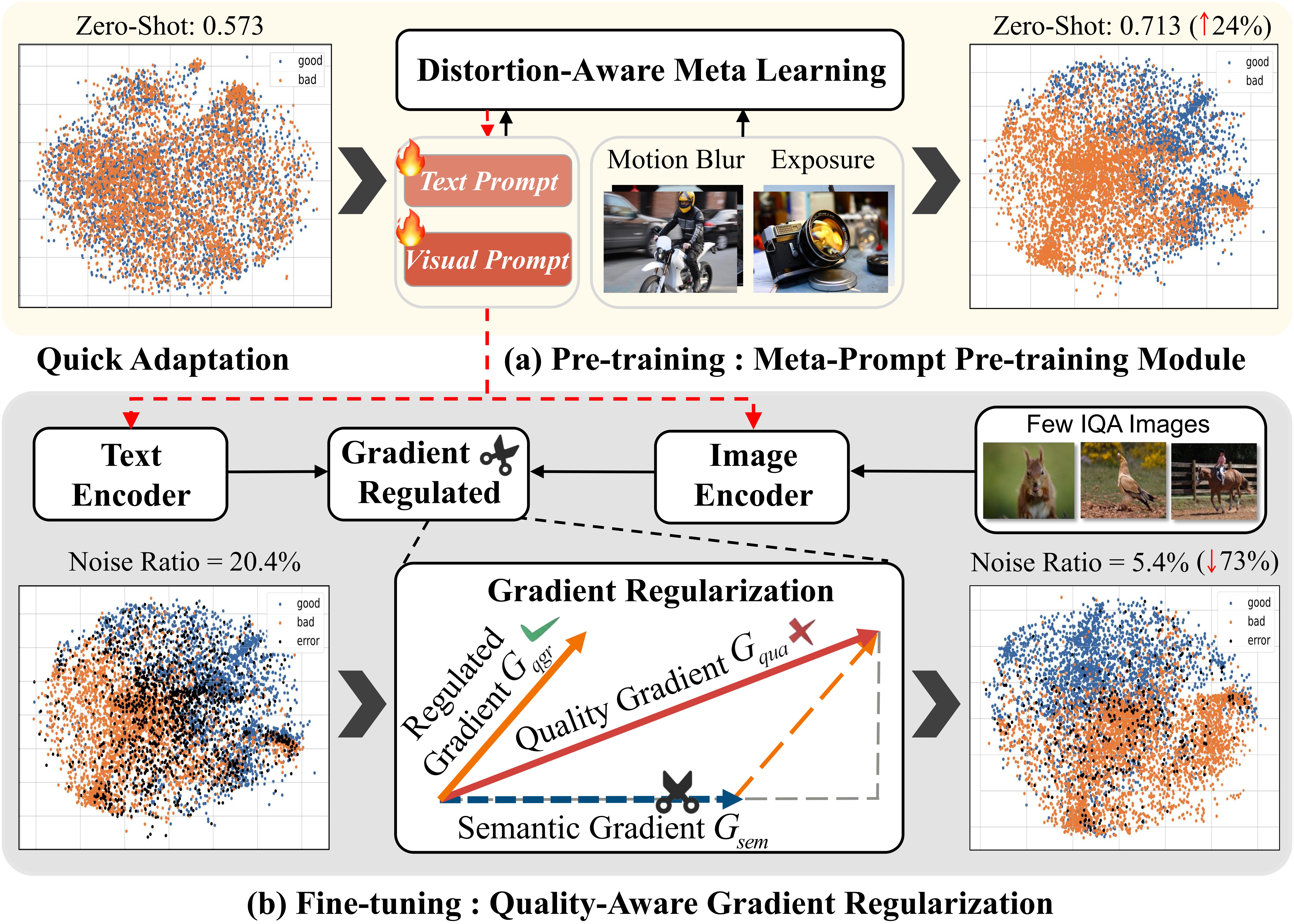}
  \caption{
 Intuitive diagram of GRMP-IQA. (a) demonstrates using meta-learning for efficient soft prompt initialization with quality prior that enhances zero-shot generalization~(\textcolor{black}{\bm{$\uparrow24\%$}}), enabling the CLIP to adapt to BIQA tasks effectively. 
 (b) illustrates gradient regularization during fine-tuning, which clips quality gradients aligned with semantic bias to guide the model toward quality-relevant features and reduce overly reliance on semantic content. In the t-SNE visualization, black dots represent noise samples with high semantic confidence ($\text{confidence} \boldsymbol{> 0.8}$) but incorrect quality predictions.
 The noticeable decrease~(\textcolor{black}{\bm{$\downarrow73\%$}}) in black dots highlights reduced semantic noise interference in IQA tasks.}
  \label{fig1}
    \vspace{-10pt}
\end{figure}
With the rise of the mobile internet era, the focus on computer vision has transitioned from initial concerns with compression and image processing~\cite{sheikh2006statistical} to handling user-generated content like smartphone photos and videos~\cite{tu2021ugc,gao2025exploring, qu2024exploring}, and lately to AI-generated content~\cite{li2023agiqa,gao2024meta}. This evolution has sharply increased demand for effective Blind Image Quality Assessment (BIQA) techniques, highlighting the importance of developing methodologies that can adeptly evaluate image quality without reference images.
Data-driven BIQA models~\cite{song2023active,yang2022maniqa,qu2024bringing} based on deep neural networks have made significant progress in recent years. However, the quality scores for distorted images are often measured using the Mean Opinion Score (MOS), which is the average of multiple ratings (sometimes up to 120). As a result, acquiring a sufficient number of IQA training samples is quite labor-intensive and cost-expensive.

To mitigate this challenge, recent BIQA approaches~\cite{DEIQT,loda,reiqa,QPT,srinath2024learning} have leveraged large-scale image datasets through transfer learning or self-supervised learning. While these approaches improve performance, they often rely on computationally intensive pretraining or require domain-specific pretext tasks, limiting scalability.
On the other hand, the vision-language model CLIP~\cite{radford2021learning} has shown remarkable generalization across various downstream tasks, offering a potential alternative. Specifically, CLIP-IQA~\cite{clipiqa} demonstrates promising zero-shot performance in IQA by leveraging simple handcrafted prompts (e.g., ``Good/Bad photo''). However, a significant performance gap remains compared to fully supervised methods~\cite{srinath2024learning,q-align}, highlighting the need to leverage limited labeled data to further close this performance gap.
To bridge this gap, we introduce the concept of \textbf{Few-Shot IQA}, which aims to enable models to quickly generalize to new distortion types and unseen scenarios with only a small amount of labeled data~(e.g., 50–200 training images).
A natural way to adapt CLIP to downstream IQA tasks under this setup is through prompt tuning~\cite{zhou2022learning,zhu2023prompt,jia2022visual}. Prompt tuning techniques learn soft textual prompts (continuous embeddings) from a small labeled dataset while keeping the pretrained model parameters frozen, offering a lightweight and efficient solution.

However, as shown in Tab.~\ref{data-efficient}, the direct application of the prompt tuning method CoOp~\cite{zhou2022learning} to IQA yields minimal improvement over simple linear probing on the CLIP image encoder, falling short of the expected benefits seen in high-level vision tasks.
We attribute this limitation to a fundamental mismatch between CLIP's pretraining objectives and the specific requirements of IQA tasks. CLIP is pretrained to align text and images in a shared semantic space, with a focus on high-level semantics such as objects and scenes. In contrast, IQA demands fine-grained sensitivity to low-level distortion cues—an ability in which CLIP is relatively deficient~\cite{LIQE,luo2023}. This discrepancy introduces two key challenges when applying prompt tuning to IQA tasks:

\noindent\textbf{(1)} \textbf{Sensitivity to Initialization:}  
During prompt tuning, with CLIP’s parameters frozen and limited sensitivity to distortions, the model depends heavily on the learned prompts for quality assessment. Different prompt initializations lead to distinct optimization directions~\cite{lester2021power}, which may steer optimization toward features less relevant to quality, causing significant performance variability. As shown in Fig.~\ref{fig2}, the average SRCC fluctuates considerably across different random initializations, requiring careful tuning for each IQA scenario and limiting the model’s adaptability.

\noindent\textbf{(2)} \textbf{Overfitting from Semantic Bias:}  
Due to CLIP’s inherent reliance on semantic information, fine-tuning with a small sample size often leads to overfitting, where the model learns spurious correlations (e.g., over-reliance on semantic features to infer quality~\cite{qfm-iqm,babu2023no}) rather than true image quality indicators, which degrades IQA performance. As illustrated in Fig.~\ref{fig1}, when the CLIP encoder is fine-tuned on a limited dataset, the latent space contains many instances (black dots) that confidently and accurately predict image semantics (with confidence levels above 0.8) but fail to reliably assess image quality.

To overcome these limitations, we propose Gradient-Regulated Meta-Prompt Learning for IQA (GRMP-IQA), a framework with two key components:
\textbf{(i) Meta Prompt Pre-training~(MPP)} mitigates sensitivity to prompt initialization by incorporating generalized distortion priors into the initialization of soft prompts, which enhances their adaptability to new IQA scenarios. By designing meta-training tasks with a well-annotated dataset containing diverse distortions (e.g., overexposure, blur) and optimizing prompts through bi-level gradient descent, we enable the model to learn shared meta-knowledge of quality representations across different distortions, thereby improving generalization across IQA tasks.  
\textbf{(ii) Quality-Aware Gradient Regularization (QGR)} addresses overfitting to semantic information by regulating gradient updates during fine-tuning, which balances the interplay between the quality knowledge gradient \( \boldsymbol{G}_{\text{qua}} \) and the semantic knowledge gradient \( \boldsymbol{G}_{\text{sem}} \). By clipping \( \boldsymbol{G}_{\text{qua}} \) along \( \boldsymbol{G}_{\text{sem}} \) when their directions are overly aligned, this module suppresses semantic bias and guides the model to focus on distortion-related features, improving its ability to capture low-level quality details.
Additionally, fine-tuning only CLIP's text branch can misalign quality perception between the image and text branches, limiting generalization~\cite{khattak2023maple}. To address this, we integrate text prompt tuning (CoOp~\cite{zhou2022learning}) and visual prompt tuning (VPT~\cite{jia2022visual}) by jointly meta-learning initializations for both, ensuring complementary optimization to better adapt CLIP to new IQA scenarios~(Tab.~\ref{ablation_mera}).

Our contributions can be summarized as follows:
\begin{figure}[t]
  \centering
    \includegraphics[width=0.47\textwidth]{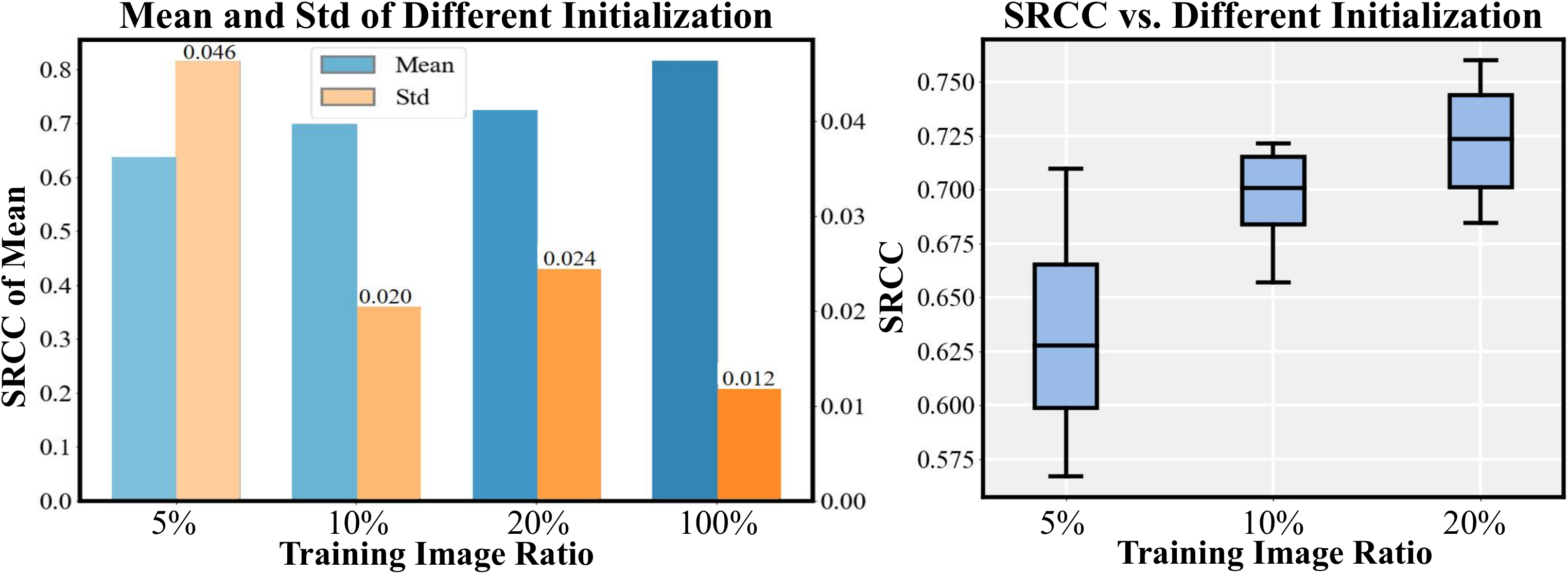}
  \caption{
  The fine-tuning results on the LIVEC dataset using CoOp~\cite{zhou2022learning} show that prompt tuning accuracy is highly sensitive to initial random initialization of the prompt, especially with limited training data, underscoring its critical role under data constraints. }
  \label{fig2}
  \vspace{-10pt}
\end{figure}  
\begin{itemize}
    \item We propose a Meta Prompt Pre-training method that organizes meta tasks based on image distortion types and optimizes soft textual and visual prompts to gain shared quality meta-knowledge, enabling CLIP to rapidly adapt across various IQA scenarios.
    \item We develop a novel Quality-Aware Gradient Regularization method that clips gradients aligned with semantic directions to balance semantic and quality information during fine-tuning, ensuring the model prioritizes image quality while still integrating relevant semantic context.
    \item Extensive empirical results confirm that our approach is both effective and efficient. Notably, with just 200 data samples, our method is competitive with SOTA models on the LIVEC dataset (using 20\% of the training data).
\end{itemize}

\section{Related Work}
\subsection{Deep Learning Based BIQA methods}
Early CNN-based IQA methods followed standard pre-training and fine-tuning pipelines~\cite{zhang2018blind, he2016deep}, while meta-learning approaches like MetaIQA~\cite{zhu2020metaiqa} improved adaptability from synthetic to real-world images. However, CNNs struggle with non-local features, a gap addressed by ViT-based models~\cite{TReS,musiq,CSIQA,li2025distilling}. Recent methods like LIQE~\cite{LIQE} leverage CLIP’s multi-task learning, enhancing IQA through supervised fine-tuning across datasets. While effective, these methods rely heavily on extensive annotations, making them costly and time-intensive.
In contrast, limited-data BIQA remains underexplored. CLIP-IQA\cite{clipiqa} shows promise in zero-shot settings but has performance limits, highlighting the need for better use of limited labeled data. Methods like DEIQT\cite{DEIQT} and LoDa\cite{loda} demonstrate that fine-tuning pre-trained ViTs reduces annotation needs but still underperform in few-shot scenarios. Self-supervised approaches\cite{reiqa,QPT,srinath2024learning} ease data constraints through tailored pre-training tasks but carry high computational costs. Additionally, recent studies~\cite{qfm-iqm,babu2023no} explore disentangling semantic and distortion-related content for IQA—yet none fully address the dual challenges of data efficiency and computational cost.

To bridge these gaps, our approach combines CLIP’s pre-trained semantic knowledge with distortion-specific fine-tuning. By using prompt tuning to cut pre-training costs and a parameter-free regularization strategy (leveraging high- and low-level gradient correlations), it optimizes IQA more effectively with fewer labeled samples.
\subsection{Prompt Tuning}
Given the limited labeled data during training, prompt tuning is an effective approach to adapt vision-language pre-trained models for few-shot learning tasks~\cite{fu2021meta,zhang2024dept,chartmoe}. CoOp~\cite{zhou2022learning} optimizes prompt vectors in CLIP’s language branch for task adaptation but struggles with generalization to novel tasks. To address this, CoCoOp~\cite{zhou2022conditional} introduces a lightweight meta-network to generate input-conditioned tokens, improving adaptability. Beyond textual prompts, visual prompts have also been explored for task adaptation~\cite{jia2022visual}. Methods like MaPLe~\cite{khattak2023maple} and PromptSRC~\cite{khattak2023self} incorporate trainable prompts into both language and visual branches, achieving significant performance gains across base and novel tasks. Recent studies~\cite{park2024prompt,li2023gradient} further construct meta-tasks based on semantic categories to initialize well-structured soft prompts, mitigating generalization decline in high-level tasks.
Unlike these approaches that focus on high-level tasks, our proposed prompt tuning method is specifically designed for IQA, leveraging low-level distortion information to construct meta-tasks. This enables pre-trained prompts to be distortion-sensitive and effectively capture shared quality knowledge.

\begin{figure*}[t]
  \centering
\includegraphics[width=1\textwidth]{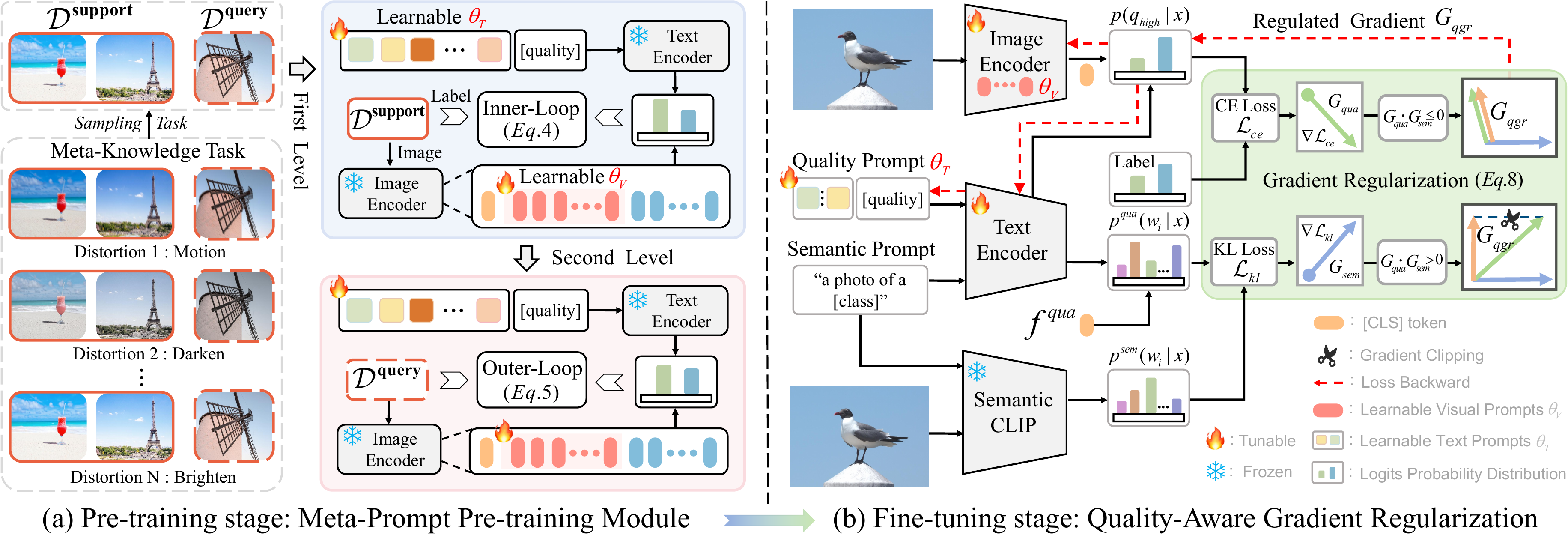}
  \caption{The overview of our GRMP-IQA. The core modules are the Meta-Prompt Pre-training and Quality-Aware Gradient Regularization, corresponding to two training processes. At the pre-training stage, we establish a distortion-specific meta-knowledge task for BIQA tasks, and the bi-level gradient descent is utilized to train visual-text meta-prompt $\textbf{[}\boldsymbol{\theta_T},\boldsymbol{\theta_V}\textbf{]}$~(Sec.~\ref{meta-learning}). These optimized prompts then serve as initial settings to efficiently adapt the CLIP model to IQA tasks with limited labels. At the few-shot fine-tuning stage, we first predict the class probability distributions $\boldsymbol{p^{sem}(w_i|x)}$ and $\boldsymbol{p^{qua}(w_i|x)}$ using the semantic and fine-tuned CLIP models, respectively. We then adjust IQA task gradients $\boldsymbol{G_{\text{qua}}}$ by clipping gradients aligned with semantic task gradients $\boldsymbol{G_{\text{sem}}}$ to generate refined gradient $\boldsymbol{G_{\text{qgr}}}$ for backward updates (Sec.~\ref{qgr}), mitigating the influence of semantic noise on quality predictions.
}
\vspace{-10pt}
\label{framework}
\end{figure*}

\section{Methodology}
\subsection{Overview}
In this paper, we propose the \textit{Gradient-Regulated Meta-Prompt Image Quality Assessment} (GRMP-IQA), which aims to adapt the CLIP for BIQA tasks with a few training samples. As depicted in Fig.~\ref{framework}, GRMP-IQA consists of two primary module: (i) the \textit{Meta-Prompt Pre-training Module} (MPP) and (ii) the \textit{Quality-Aware Gradient Regularization} (QGR). The MPP module pre-trains visual-text prompts to acquire shared meta-knowledge on distortions, enabling quick adaptation to various IQA scenarios. The QGR module plays a key role in fine-tuning by adjusting gradient updates to prevent overfitting to semantic content. 

\noindent\textbf{Pre-training stage~(Sec.~\ref{meta-learning}).} We randomly sample a mini-batch from distortion meta-tasks, partitioning it into a support set $\mathcal{D}^\text{support}$ and a query set $\mathcal{D}^\text{query}$. A bi-level gradient descent method progresses from the Inner-Loop on $\mathcal{D}^\text{support}$ to the Outer-Loop on $\mathcal{D}^\text{query}$ to optimize the learnable visual-textual prompts $\textbf{[}\boldsymbol{\theta_T},\boldsymbol{\theta_V}\textbf{]}$. These prompts are then used as initial weights for following fine-tuning. 

\noindent\textbf{Fine-tuning stage~(Sec.~\ref{qgr}).} Given an input image $\boldsymbol x$ and a hard semantic prompt $\boldsymbol{w}_i =$ ``a photo of a [class]'', we predict the semantic class distributions $\boldsymbol{p^{sem}(w_i|x)}$ and $\boldsymbol{p^{qua}(w_i|x)}$ using the original semantic CLIP model and the fine-tuned CLIP model, respectively. The gradient of the KL divergence loss $\boldsymbol{\nabla \mathcal{L}_{kl}}$ between these distributions is computed as the general semantic direction $\boldsymbol{G}_{\text{sem}}$. Simultaneously, using soft quality prompt $\boldsymbol{\theta_T}$ and image $\boldsymbol x$, we compute the gradient $\boldsymbol{\nabla \mathcal{L}_{ce}}$ of the quality loss as the quality direction $\boldsymbol{G}_{\text{qua}}$. To refine the optimization direction of the IQA task, we adjust $\boldsymbol{G}_{\text{qua}}$ into $\boldsymbol{G}_{\text{qgr}}$ by clipping its components aligned with $\boldsymbol{G}_{\text{sem}}$, ensuring the model focuses on image quality while reducing the impact of semantic noise.

\subsection{Visual-Text Meta-Prompt}
\noindent \textbf{Visual Meta-Prompt.}
We utilize {Deep Prompt Tuning (DPT)}~\cite{zhou2022learning} as our {Visual Meta-Prompt}, with learnable parameters $\boldsymbol{\theta _V}$. The input embedding for the $l$-th layer's self-attention module in the ViT-based image encoder is denoted as $\{\boldsymbol{f}^{l}, \boldsymbol{H}^{l}\}$, where $\boldsymbol{f}^{l}$ represents the classification (CLS) token, and $\boldsymbol{H}^{l} = \{\boldsymbol{h}_{1}^{l}, \boldsymbol{h}_{2}^{l}, \ldots, \boldsymbol{h}_{N}^{l}\}$ denotes the image patch embeddings. A learnable token $\boldsymbol{P}^{l}$ is appended to the token sequence in each ViT layer, and the {Multi-Head Self-Attention (MHSA)} module processes the tokens as:
\begin{equation}
[\boldsymbol{f}^l, \_, \boldsymbol{H}^l] = \text{Layer}^{l}([\boldsymbol{f}^{{l-1}}, \boldsymbol{P}^{l-1}, \boldsymbol{H}^{l-1}]),
\end{equation}
where the output of $\boldsymbol{P}^{l}$ is discarded and not passed to the next layer, serving only as a set of learnable parameters.

\noindent \textbf{Text Meta-Prompt.}
We adopt CoOp \cite{zhou2022learning} as our textual meta-prompt, with learnable parameters $\boldsymbol{\theta _T}$.  
For each quality class~$i$, we construct a learnable quality prompt defined as $\boldsymbol{q}_{i} = \{\boldsymbol{v_1}, \boldsymbol{v_2}, \ldots, \boldsymbol{v_M}, [\text{quality}]\}$. Here, $[\text{quality}] \in \{\text{``high quality''}, \text{``low quality''}\}$ serves as a categorical quality marker, and $\{\boldsymbol{v}_{m}\}_{m=1}^{M}$ denotes a set of $M$ learnable context vectors (with $M=4$ in our work).
\subsection{Meta-Prompt Pre-training Module}~\label{meta-learning}
During meta-learning, we aim to optimize the learnable visual-text prompt $\boldsymbol{\theta = [{\theta _T},{\theta _V}]}$. For an input image $\boldsymbol{x}$, the textual encoder $\boldsymbol{g(\cdot)}$ processes the quality prompts $\boldsymbol{q}_{i}$, while the visual encoder extracts the feature vector $\boldsymbol{f}$. The probability of predicting high quality is:
\begin{equation}
p(\boldsymbol{q}_\text{high}|\boldsymbol{x}) = \frac{\exp(\langle \boldsymbol{g}(\boldsymbol{q}_\text{high}), \boldsymbol{f} \rangle / \boldsymbol{\tau})}{\exp(\langle \boldsymbol{g}(\boldsymbol{q}_\text{high}), \boldsymbol{f} \rangle / \boldsymbol{\tau}) + \exp(\langle \boldsymbol{g}(\boldsymbol{q}_\text{low}), \boldsymbol{f} \rangle / \boldsymbol{\tau})},
\end{equation}
where $\boldsymbol{q}_\text{high}$ and $\boldsymbol{q}_\text{low}$ denote high quality and low quality category prompt, \({p(\boldsymbol{q}_\text{high}|\boldsymbol{x})}\) represents the estimated probability that the quality of image \(\boldsymbol{x}\) is ``high quality'', $\boldsymbol{\tau}$ is a temperature parameter learned by CLIP, and $\boldsymbol{\left\langle, \right\rangle}$ denotes cosine similarity.
The labeled quality scores are then rescaled to 0-1, denoted as $y$, and the loss function is calculated as:
\begin{equation}
\mathcal{L} =  - (y\log (p(q_\text{high}|\boldsymbol{x})) + (1 - y)\log (1 - p(q_\text{high}|\boldsymbol{x}))).
\label{loss}
\end{equation}

\noindent \textbf{Constructing Distortion Meta-Knowledge Task.}
As noted by \cite{ma2019blind,zhu2020metaiqa}, the ability to detect various types of image distortions is crucial for developing BIQA models with strong generalization capabilities across diverse scenarios. Moreover, the efficacy of prompt tuning is significantly dependent on the initial configuration of the prompts. This initial setup greatly affects the CLIP vision-language model's ability to swiftly adapt to different IQA scenarios. Drawing inspiration from the ``learn to learn'' ethos inherent in the deep meta-learning paradigm \cite{vanschoren2018meta,zhu2020metaiqa}, we propose an optimization-based method for effectively pre-training visual-textual prompts. These prompts incorporate shared quality insights from various image distortions, thereby enhancing CLIP's swift adaptability to IQA tasks~\cite{zhu2020metaiqa}. 
To investigate the general rules of image distortion, we first constructed a $K_t$-way distortion-specific image quality prediction task, denoted as $\mathcal{T}_t$. Then, it is used to build the meta-training set as \(\mathcal{D}^{\text{meta}} = \left\{\mathcal{D}_{\mathcal{T}_t}^\text{support}, \mathcal{D}_{\mathcal{T}_t}^\text{query}\right\}_{t=1}^T\). Here, \(\mathcal{D}_{\mathcal{T}_t}^\text{support}\) and \(\mathcal{D}_{\mathcal{T}_t}^\text{query}\) represent the support and query sets for each task, respectively, with \(T\) representing the total number of tasks. 
To simulate the process of prompt generalization to different distortions in BIQA scenarios, we randomly sample \(k\) tasks as a mini-batch from the meta-training set, where \(1 \leq k \leq T\), to perform bi-level gradient optimization.

\noindent \textbf{Distortion-Aware Meta-prompt Learning.} 
Our approach employs a bi-level gradient descent technique to bridge the learning process from the support to the query set. Specifically, it mainly consists of two optimization steps. In the Inner-Loop (the first level), we compute the gradients of the prompt parameters using the support set and apply the first update. In the Outer-Loop (the second level), we assess the performance of the updated model on the query set and optimize the parameters again. This bi-level structure distills shared quality priors by training on a number of NR-IQA tasks with known distortions, enabling the meta-prompt to generalize rapidly across diverse new BIQA scenarios.

\noindent\textbf{Inner-Loop.}
The objective of the Inner-Loop stage is to adapt the meta-prompt, denoted by $\textbf{[}\boldsymbol{\theta_T},\boldsymbol{\theta_V}\textbf{]}$, to the $t^{th}$ support set ${\mathcal{D}}_{{\mathcal{T}_t}}^{\text{support}}$ within the mini-batch. During the first level of updates, we determine the loss $\mathcal{L}(\boldsymbol{\theta}, \mathcal{D}_{\mathcal{T}_t}^{\text{support}})$, following which the model parameters are updated on the support set using the inner learning rate $\alpha$, as specified by:
\begin{equation}
\boldsymbol{\theta}'_t = \boldsymbol{\theta} - \alpha \nabla_{\boldsymbol{\theta}} \mathcal{L}(\boldsymbol{\theta}, \mathcal{D}_{\mathcal{T}_t}^{\text{support}}),
\end{equation}

\noindent\textbf{Outer-Loop.} In a similar vein, the second level of updates adjusts the parameters $\boldsymbol{\theta}'_t$ based on the query set ${\mathcal{D}_{\mathcal{T}_t}^\text{query}}$:
\begin{equation}
\boldsymbol{\theta}_t \leftarrow \boldsymbol{\theta}'_t - \alpha \nabla_{\boldsymbol{\theta}'_t} \mathcal{L}(\boldsymbol{\theta}'_t, \mathcal{D}_{\mathcal{T}_t}^\text{query}).
\end{equation}
For a mini-batch of meta-tasks, this process culminates in aggregating gradients from all tasks to update the final model parameters, following the update rule:
\begin{equation}
\boldsymbol{\theta} \leftarrow \boldsymbol{\theta} - \beta \sum_{t=1}^{k} \nabla_{\boldsymbol{\theta}} \mathcal{L}(\boldsymbol{\theta}'_t, \mathcal{D}_{\mathcal{T}_t}^{\text{query}}),
\end{equation}
where $\beta$ represents the outer learning rate. Meta-learning effectively trains the learnable prompts $\boldsymbol{\theta} = \textbf{[}\boldsymbol{\theta_T},\boldsymbol{\theta_V}\textbf{]}$, ensuring their generalization across various image distortions.
\subsection{Quality-Aware Gradient Regularization}\label{qgr}
We denote the original semantic CLIP model as $\boldsymbol{V}^{\text{sem}}$ and the model obtained from the Meta-Prompt Pre-training Module, as $\boldsymbol{V}^{\text{qua}}$. The gradient for the IQA task is the quality gradient $\boldsymbol{G}_{\text{qua}}$, while the gradient for the semantic direction is the semantic gradient $\boldsymbol{G}_{\text{sem}}$.
Recent studies \cite{QPT,babu2023no} have shown that IQA tasks can be misaligned with the high-level semantic representations of upstream tasks, leading to overfitting and reduced generalization.  To prevent semantic overfitting, our Quality-aware Gradient Regularization (QGR) follows three key steps: derive a semantic direction $\boldsymbol{G}_{\text{sem}}$, compute the quality optimization direction $\boldsymbol{G}_{\text{qua}}$, and balance gradients based on their directional relationship.

Specifically, to derive a general semantic direction $\boldsymbol{G}_{\text{sem}}$, we first design a hard prompt following~\cite{ying2020patches,LIQE}: $\boldsymbol{w}_i = \text{``a photo of a [class]''}$, where [class] represents one of nine categories: {``animal'', ``cityscape'', ``human'', ``indoor'', ``landscape'', ``night'', ``plant'', ``still-life'', ``others''}. This prompt $\boldsymbol{w}_i$ is input into the text encoders of $\boldsymbol{V}^{\text{sem}}$ and $\boldsymbol{V}^{\text{qua}}$, generating text features that align with image features $\boldsymbol{f}^{qua}$ from their respective visual encoders. As a result, we obtain zero-shot semantic prediction probabilities $\boldsymbol{p}^{\text{sem}}(\boldsymbol{w}_i|\boldsymbol{x})$ and $\boldsymbol{p}^{\text{qua}}(\boldsymbol{w}_i|\boldsymbol{x})$. To measure the semantic alignment between $\boldsymbol{V}^{\text{qua}}$ and $\boldsymbol{V}^{\text{sem}}$, we compute the KL divergence:
\begin{equation}
    \mathcal{L}_{\text{kl}}(\boldsymbol{V}^{\text{qua}}) = - \sum_i \boldsymbol{p}^{\text{sem}}(\boldsymbol{w}_i|\boldsymbol{x}) \log \frac{\boldsymbol{p}^{\text{qua}}(\boldsymbol{w}_i|\boldsymbol{x})}{\boldsymbol{p}^{\text{sem}}(\boldsymbol{w}_i|\boldsymbol{x})}.
\end{equation}
The gradient of $\mathcal{L}_{\text{kl}}(\boldsymbol{V}^{\text{qua}})$, denoted as $\boldsymbol{G}_{\text{sem}}$, represents the general semantic optimization direction.
Similarly, the quality direction $\boldsymbol{G}_{\text{qua}}$ is obtained by computing the gradient of the cross-entropy loss $\mathcal{L}_{\text{ce}}(\boldsymbol{V}^{\text{qua}})$, which compares the predicted probability $\boldsymbol{p}(\boldsymbol{q}_{\text{high}}|\boldsymbol{x})$ with the ground truth $y$, as shown in Eq.~\ref{loss}.
Finally, to balance the gradients between quality $\boldsymbol{G}_{\text{qua}}$ and semantics $\boldsymbol{G}_{\text{sem}}$ within a shared representation space, we analyze their relationship under two cases:

\noindent \textbf{(1)} If the angle between $\boldsymbol{G}_{\text{qua}}$ and $\boldsymbol{G}_{\text{sem}}$ is less than 90 degrees, it indicates that the optimization directions for IQA quality knowledge and general semantic knowledge are consistent. In such cases, we clip the $\boldsymbol{G}_{\text{qua}}$ along its component $\boldsymbol{G_{\parallel}}$ that parallel to the semantic direction $\boldsymbol{G}_{\text{sem}}$, to modulate the model's original quality optimization path, preventing overfitting to semantic correlations.

\noindent \textbf{(2)} Conversely, if the angle between two areas of knowledge is more than 90 degrees, it means they are heading in different directions during improvement. In such situations, we don't adjust the quality gradient $\boldsymbol{G}_{\text{qua}}$, allowing the model to learn quality-aware features according to the original optimization direction.
In conclusion, our gradient regularization strategy is mathematically formulated as:
\begin{equation}
\boldsymbol{G}_{\text{qgr}} = 
\begin{cases} 
\boldsymbol{G}_{\text{qua}} & \text{if } \boldsymbol{G}_{\text{qua}} \cdot \boldsymbol{G}_{\text{sem}} \leq 0, \\
\boldsymbol{G}_{\text{qua}} - \lambda \frac{\boldsymbol{G}_{\text{qua}} \cdot \boldsymbol{G}_{\text{sem}}}{\|\boldsymbol{G}_{\text{sem}}\|^2} \boldsymbol{G}_{\text{sem}} & \text{otherwise},
\end{cases}
\label{eq_qgr}
\end{equation}
Here, $\lambda$ is introduced to generalize the formulation, providing flexibility in controlling the influence of general knowledge. Specifically, $\lambda = 1$ projects $\boldsymbol{G}_{\text{qua}}$ onto the orthogonal direction of $\boldsymbol{G}_{\text{sem}}$, while $\lambda = 0$ reduces QGR to CoOp~\cite{zhou2022learning}.
\begin{table*}[t]
\centering
\renewcommand{\arraystretch}{1.0}
\setlength{\tabcolsep}{4pt} 
\resizebox{1\textwidth}{!}{
\begin{tabular}{c|ccc|ccc|ccc|ccc|ccc} 
\toprule
Method         & \multicolumn{3}{c|}{LIVEC} & \multicolumn{3}{c|}{KonIQ} & \multicolumn{3}{c|}{CSIQ} & \multicolumn{3}{c|}{LIVE} & \multicolumn{3}{c}{PIPAL}  \\ 
\midrule
Labels         & 50    & 100   & 200        & 50    & 100   & 200        & 50    & 100   & 200       & 50    & 100   & 200       & 50    & 100   & 200         \\ 
\midrule
HyperIQA~\cite{hypernet}  & 0.648 & 0.725 & 0.790       & 0.615 & 0.710  & 0.776      & 0.790  & 0.824 & 0.909     & 0.892 & 0.912 & 0.929     & 0.102 & 0.302 & 0.379       \\
MetaIQA~\cite{zhu2020metaiqa} & 0.604 & 0.626 & 0.669      & 0.618  &0.620 & 0.660      & 0.784 & 0.849 & 0.894      & 0.840 & 0.880 & 0.919   & 0.332 & 0.348 & 0.371\\ 
DEIQT~\cite{DEIQT}     & 0.667 & 0.718 & 0.812      & 0.638 & 0.682 & 0.754      & 0.821 & 0.891 & 0.941     & 0.920  & \underline{0.942} & 0.955     & \underline{0.396} & 0.410  & 0.436       \\
MANIQA~\cite{yang2022maniqa}    & 0.642 & 0.769 & 0.797      & 0.652 & 0.755 & 0.810       & 0.794 & 0.847 & 0.874     & 0.909 & 0.928 & \underline{0.957}     & 0.136 & 0.361 & 0.470       \\
CONTRIQUE~\cite{madhusudana2022image} & 0.695 & 0.729 & 0.761      & 0.733 & 0.794 & 0.821      & 0.840  & \textbf{0.926} & 0.940      & 0.891 & 0.922 & 0.943     & 0.379 & 0.437 & 0.488       \\
Re-IQA~\cite{reiqa}    & 0.591 & 0.621 & 0.701      & 0.685 & 0.723 & 0.754      & \underline{0.893} & 0.907 & 0.923     & 0.884 & 0.894 & 0.929     & 0.280  & 0.350  & 0.431       \\
\midrule
 CLIP w/ Linear Probe~\cite{radford2021learning}      & 0.664 & 0.721 & 0.733      & 0.736 & 0.770  & 0.782      & 0.841 & 0.892 & 0.941     & 0.896 & 0.923 & 0.941     & 0.254 & 0.303 & 0.368       \\
 $\text{CLIP w/ CoOp~\cite{clipiqa}}^{\dagger}$      & 0.695 & 0.738 & 0.746      & 0.692 & 0.743  & 0.762      & - & - & -     & - & - & -     & - & - & -       \\
 LIQE~\cite{LIQE}      & 0.691 & 0.769 & 0.810       & 0.759 & 0.801 & 0.832      & 0.838 & 0.891 & 0.924     & 0.904 & 0.934 & 0.948     & -     & -     & -           \\
GRepQ~\cite{srinath2024learning} & \underline{0.760} & \underline{0.791} & \underline{0.822} & \underline{0.812} & \underline{0.836} & \underline{0.855} & 0.878 & 0.914 & \underline{0.941} & \underline{0.926} & {0.937} & {0.953} & $0.390^{\dagger}$ & $\underline{0.450}^{\dagger}$ & $\underline{0.498}^{\dagger}$ \\
\midrule
\rowcolor{cyan!10} GRMP-IQA (Ours) & \textbf{0.836} & \textbf{0.857} & \textbf{0.875} & \textbf{0.853} & \textbf{0.872} & \textbf{0.883} & \textbf{0.893} & \underline{0.917} & \textbf{0.941} & \textbf{0.932} & \textbf{0.943} & \textbf{0.968} & \textbf{0.474} & \textbf{0.512} & \textbf{0.546} \\
\bottomrule
    \end{tabular}}
  \caption{SRCC performance comparison of our method with other IQA methods trained on limited labels. Bold indicates the best results, underlined marks the second-best, and the fifth-to-last through second-to-last lines show the CLIP-based IQA. $\dagger$ is our reproduction.
}  
\label{data-efficient}
\vspace{-5pt}
\end{table*}
\begin{table*}[t]
\renewcommand{\arraystretch}{0.95}
\resizebox{1\textwidth}{!}{
  \centering
    \begin{tabular}{lcccccccccccc}
    \toprule
    & \multicolumn{2}{c}{LIVE} & \multicolumn{2}{c}{CSIQ} & \multicolumn{2}{c}{LIVEC} & \multicolumn{2}{c}{KonIQ} & \multicolumn{2}{c}{LIVEFB} & \multicolumn{2}{c}{SPAQ}\\
    \cmidrule{2-13}    Method & \multicolumn{1}{c}{PLCC} & \multicolumn{1}{c}{SRCC} & \multicolumn{1}{c}{PLCC} & \multicolumn{1}{c}{SRCC} & \multicolumn{1}{c}{PLCC} & \multicolumn{1}{c}{SRCC}& \multicolumn{1}{c}{PLCC} & \multicolumn{1}{c}{SRCC}& \multicolumn{1}{c}{PLCC} & \multicolumn{1}{c}{SRCC}& \multicolumn{1}{c}{PLCC} & \multicolumn{1}{c}{SRCC}\\
    \midrule 
     Training Ratio & {20\%} & {20\%} & {20\%} & {20\%} & {20\%} & {20\%} & {20\%} & {20\%} & {10\%} & {10\%} & {20\%} & {20\%} \\
     \midrule
    DEIQT~\cite{DEIQT} & {0.968} & {0.965} & {0.885} & {0.862} & {0.822} & {0.792} & {0.908} & {0.888} & {0.624} & {0.538} & {0.912} & {0.908} \\
    LoDa~\citep{loda} & {-} & {-} & - & - & {0.854} & {0.815} & {0.923} & 0.907 & {-} & {-} & {-} & {-} \\
    \rowcolor{cyan!10} GRMP-IQA (Ours) & {0.972} & {0.970} & {0.958} & {0.951} & {0.897} & {0.875} & {0.931} & {0.915} & {0.686} & {0.604} & {0.925} & {0.920} \\
    \midrule
     Training Ratio & {80\%} & {80\%} & {80\%} & {80\%} & {80\%} & {80\%} & {80\%} & {80\%} & {80\%} & {80\%} & {80\%} & {80\%} \\
     \midrule
    MetaIQA~\cite{zhu2020metaiqa} & 0.959 & 0.960  & 0.908 & 0.899 & 0.802 & 0.835 & 0.856 & 0.887 & 0.507 & 0.540  & {-} & {-} \\
    CONTRIQUE~\cite{madhusudana2022image} & {0.961} & {0.960} & {0.955} & {0.942} & {0.857} & {0.845} & {0.906} & {0.894} & {0.641} & {0.580} & {0.919} & {0.914} \\
    DEIQT~\cite{DEIQT} & {0.982} & {0.980} & {0.963} & {0.946} & {0.894} & {0.875} & {0.934} & {0.921} & {0.663} & {0.571} & {0.923} & {0.919} \\
    Re-IQA~\cite{reiqa} & {0.971}  & {0.970} & {0.960} & {0.947} & {0.854} & {0.840} & 0.923 & 0.914 & \textbf{0.733} & \textbf{0.645} & {0.925} & 0.918\\ 
    LIQE~\citep{LIQE} & 0.951 & 0.970 & 0.939 & 0.936 &  {0.910} & \textbf{0.904} & 0.908 & 0.919 & - & - & {-} & {-} \\
    CLIP-IQA+~\citep{clipiqa} & - & - & - & - &  {0.832} & {0.805} & 0.909 & 0.895 & 0.593 & 0.575 & {0.866} & {0.864} \\
    CDINet~\cite{zheng2024cdinet} & {0.975} & {0.977} & {0.960} & {0.952} & {0.880} & 0.865& 0.928 & 0.916 & \textbf{-} & \textbf{-} & 0.922 & 0.919 \\
    QFM-IQM~\cite{qfm-iqm} & \textbf{0.983} & \textbf{0.981} & \textbf{0.965} & \textbf{0.954} & \textbf{0.913} & {0.891} & {0.936} & {0.922} & {0.667} & {0.567} & {0.924} & {0.920} \\
    LoDa~\citep{loda} & 0.979 & 0.975 & - & - & {0.899} & {0.876} & \textbf{0.944} & \textbf{0.932} & {0.679} & {0.578} & \textbf{0.928} & \textbf{0.925} \\
    \midrule 
    \rowcolor{cyan!10} GRMP-IQA (Ours) & \textbf{0.983} & \textbf{0.981} & \textbf{0.974} & \textbf{0.968} & \textbf{0.916} & \textbf{0.897} & \textbf{0.945} & \textbf{0.934} & \textbf{0.704} & \textbf{0.616} & \textbf{0.932} & \textbf{0.927} \\
    \bottomrule
    \end{tabular}}
        \caption{Performance comparison measured by medians of SRCC and PLCC, and \textbf{bold} entries indicate the top two results.} 
  \label{performance}
  \vspace{-15pt}
\end{table*}
\vspace{-15pt}
\section{Experiments}
\subsection{Datasets and Evaluation Protocols}
We conduct experiments on multiple BIQA datasets include LIVEC~\cite{ghadiyaram2015massive}, KonIQ~\cite{hosu2020koniq}, LIVEFB~\cite{ying2020patches}, and SPAQ~\cite{fang2020perceptual}, which feature authentic distortions, and PIPAL~\cite{jinjin2020pipal}, LIVE~\cite{sheikh2006statistical}, and CSIQ~\cite{larson2010most}, which feature synthetic distortions.
LIVEC contains 1,162 mobile device images, SPAQ includes 11,125 photos from 66 smartphones, KonIQ has 10,073 images from open sources, and LIVEFB is the largest real-world dataset with 39,810 images. For synthetic distortions, LIVE and CSIQ contain 779 and 866 images with 5 and 6 types of distortions. PIPAL, a challenging dataset, includes 23,200 images with 40 types of distortions, including GAN-generated artifacts.
We use Spearman's Rank Correlation Coefficient (SRCC) and the Pearson Linear Correlation Coefficient (PLCC) as metrics to quantify the monotonousness and accuracy of predictions.
\vspace{-10pt}
\subsection{Implementation Details and Setups}
We build our model on CLIP-B/16~\cite{radford2021learning}. During pre-training, we optimize the visual-text prompt, while in fine-tuning, only the \textbf{last four blocks} of the image and text encoders are trained. For competing models, we use public results or re-train them under our setup. Each dataset is split into 80\%/20\% for training and testing, with splits based on reference images to ensure content independence in datasets with synthetic distortions. To ensure robustness, we report the median performance across ten random splits. All experiments are run on four NVIDIA RTX3090 GPUs.

\noindent \textbf{Meta-Prompt Pre-training.}
We pre-train on TID2013~\cite{ponomarenko2015image} and KADID-10K~\cite{lin2019kadid} dataset, which contain extensive distortion information. We set the learning rates $\alpha$ and $\beta$ to 1e-4 and 1e-2, and train for 50 epochs using Adam~\cite{kingma2014adam}.

\noindent \textbf{Fully Supervised Learning Setting.}
We randomly crop each input image into 10 patches of $224 \times 224$ resolution and train the model for 9 epochs using AdamW~\cite{loshchilov2017decoupled}. The learning rate is ${5 \times 10^{-6}}$, with a scheduler over 9 decay epochs. The batch size is 16 for LIVEC and 128 for KonIQ. 

\noindent \textbf{Few-Shot Learning Setting.}
In the few-shot setting, we follow the approach in a previous study~\cite{srinath2024learning} to train our GRMP-IQA model using randomly selected subsets of 50, 100, and 200 samples from the training set. The regularization weight $\lambda$ is fixed at 5, and the training hyperparameters are set according to the schedule proposed by CoOp~\cite{zhou2022learning}.

\subsection{Performance Comparison with SOTA}
Our method effectively acquires extensive image quality assessment knowledge, enabling it to provide powerful priors for various IQA scenarios. Tab.~\ref{data-efficient} and Tab.~\ref{performance} summarize the comparative results for different experimental settings.

\noindent \textbf{Few-Shot Setting.}  
Benefiting from the quality priors acquired during pre-training, our model achieves superior performance even with limited training data. As shown in Tab.~\ref{data-efficient}, our GRMP-IQA significantly outperforms the second-best model, GRepQ~\cite{srinath2024learning}, which is specifically designed for few-shot learning. Additionally, our approach exhibits clear advantages over MetaIQA, a meta-learning-based method, underscoring the effectiveness of meta-prompts for rapidly adapting CLIP to diverse IQA tasks. These results validate GRMP-IQA's strong capability in learning quality-aware feature with limited labeled samples.

\noindent \textbf{Fully Supervised Setting.}  
Tab.~\ref{performance} compares GRMP-IQA with other BIQA methods under full supervised setting, including self-training approaches like CONTRIQUE~\cite{madhusudana2022image} and Re-IQA~\cite{reiqa}, as well as CLIP-based methods such as LIQE~\cite{LIQE} and CLIP-IQA+~\cite{clipiqa}. GRMP-IQA outperforms almost all competitors on six datasets. Achieving leading performance across these datasets is particularly challenging due to the diverse range of image content and distortion types.
Notably, the proposed method achieves competitive results with the state-of-the-art (SOTA) method on several datasets while utilizing only 20\% of the training data compared to the fully supervised setting.
\begin{table}[t]
\setlength\tabcolsep{3pt}
\renewcommand{\arraystretch}{1.0}
  \centering
    \resizebox{0.47\textwidth}{!}{
    \begin{tabular}{ccccccc}
    \toprule
    Training & \multicolumn{2}{c}{  LIVEFB } & \multicolumn{1}{c}{LIVEC} & \multicolumn{1}{c}{KonIQ} & \multicolumn{1}{c}{LIVE} & \multicolumn{1}{c}{CSIQ} \\
    \midrule
    Testing & \multicolumn{1}{c}{KonIQ} & \multicolumn{1}{c}{LIVEC} & \multicolumn{1}{c}{KonIQ} & \multicolumn{1}{c}{LIVEC} & \multicolumn{1}{c}{CSIQ} & \multicolumn{1}{c}{LIVE} \\
    \midrule
    DBCNN & 0.716 & 0.724 & 0.754 & 0.755 & 0.758 & 0.877 \\
    HyperIQA & {0.758} & 0.735 & {0.772} & 0.785 & 0.744 & 0.926 \\
    TReS  & 0.713 & 0.740  & 0.733 & 0.786 & 0.761 & {-} \\
    DEIQT & 0.733 & 0.781 & 0.744 & 0.794 & 0.781 & 0.932 \\
    CLIP-IQA+ & 0.631 & 0.620 & 0.697 & 0.803 & - & -\\
    LoDa & {0.763} & \textbf{0.805} & 0.745 & {0.811} & {-} & {-} \\
    \midrule
    \rowcolor{cyan!10} GRMP-IQA  & \textbf{0.765} & {0.790} & \textbf{0.782} & \textbf{0.831} & \textbf{0.809} & \textbf{0.935} \\
    \bottomrule
    \end{tabular}}
        \caption{SRCC on the cross datasets validation. The best performances are highlighted in boldface.}
  \label{crossdata}%
    \vspace{-5pt}
\end{table}%
\begin{table}[t]
\centering
\setlength\tabcolsep{2.5pt} 
\renewcommand{\arraystretch}{1.1} 
\resizebox{0.47\textwidth}{!}{
\begin{tabular}{cc|cccc} 
\toprule
\multicolumn{2}{c|}{Component} & \multicolumn{2}{c}{LIVEC} & \multicolumn{2}{c}{KonIQ}  \\ 
\midrule
Pre-training & Meta-learning & PLCC & SRCC               & PLCC & SRCC                \\ 
\midrule
        &      & 0.825    & 0.796                  & 0.788    & 0.764                  \\
\CheckmarkBold       &      & 0.823    & 0.788                  & 0.792    & 0.761                   \\
\rowcolor{cyan!10} \CheckmarkBold       & \CheckmarkBold    & \textbf{0.858}    & \textbf{0.828}                  & \textbf{0.844}    & \textbf{0.811}                   \\
\bottomrule
\end{tabular}}
\caption{Ablation study on the effectiveness of meta-learning when trained with only 50 samples.}
    \label{ablation_meta}
    \vspace{-15pt}
\end{table}
\subsection{Generalization Capability Validation}
To evaluate the generalization capacity of GRMP-IQA, we performed cross-dataset validation experiments, where the model was trained on one dataset and tested on others without parameter adjustments. Tab.~\ref{crossdata} reports the SRCC results across five datasets. GRMP-IQA consistently outperforms state-of-the-art models in most cross-authentic scenarios, achieving notable improvements on LIVEC and KonIQ dataset. Additionally, it demonstrates strong competitiveness on synthetic datasets like LIVE and CSIQ.
\begin{table}[t]
\setlength\tabcolsep{3pt}
\centering
 \resizebox{0.47\textwidth}{!}{
\begin{tabular}{ccc|cc|cc|cc} 
    \toprule
     \multicolumn{3}{c|}{Components} & \multicolumn{2}{c|}{LIVEC} & \multicolumn{2}{c|}{KonIQ} & \multicolumn{2}{c}{PIPAL}  \\ 
\midrule
Meta & Text & Visual & PLCC  & SRCC               & PLCC  & SRCC               & PLCC  & SRCC               \\ 
    \midrule
     &      &      & 0.579 & 0.598              & 0.592 & 0.573              & 0.216 & 0.203              \\
\CheckmarkBold    &      &      & 0.639 & 0.589              & 0.556 & 0.554              & 0.367 & 0.371              \\
\CheckmarkBold    & \CheckmarkBold    &      & 0.699 & 0.689              & 0.679 & 0.609              & 0.323 & 0.312              \\
\CheckmarkBold    &      & \CheckmarkBold    & 0.776 & 0.742              & 0.736 & 0.701              & 0.357 & 0.369             \\
    & \CheckmarkBold     & \CheckmarkBold    & 0.759 & 0.709              & 0.622 & 0.592              & 0.362 & 0.396             \\
    \midrule
    \rowcolor{cyan!10}
\CheckmarkBold    & \CheckmarkBold    & \CheckmarkBold    &\textbf{0.808} & \textbf{0.770}               & \textbf{0.744} & \textbf{0.713}              & \textbf{0.410}  & \textbf{0.434}              \\
    \bottomrule
\end{tabular}}
    \caption{Ablation experiments with Meta-Prompt Pre-training component. The best performances are highlighted in boldface.}
\label{ablation_mera}
\vspace{-5pt}
\end{table}

\begin{table}[t]
\centering
 \resizebox{0.47\textwidth}{!}{
\begin{tabular}{c|c|cc|cc} 
\toprule
\multirow{2}{*}{Method} & \multirow{1}{*}{Labels} & \multicolumn{2}{c|}{50} & \multicolumn{2}{c}{100}  \\ 
\cmidrule{2-6}
                        & Dataset                          & PLCC  & SRCC            & PLCC  & SRCC             \\ 
\midrule
w/o QGR           & LIVEC                    & 0.858 & 0.828           & 0.875 & 0.848            \\
\rowcolor{cyan!10} {w/ QGR}             & LIVEC                    & \textbf{0.864} & \textbf{0.836}           & \textbf{0.883} & \textbf{0.857}            \\                        
\midrule
{w/o QGR}                                     & KonIQ                    & 0.844 & 0.811          & 0.872 & 0.840            \\
\rowcolor{cyan!10} {w/ QGR}   & KonIQ                    & \textbf{0.880} & \textbf{0.853}           & \textbf{0.896} & \textbf{0.872}            \\
\bottomrule
\end{tabular}}
\caption{Ablation experiments with QGR in few-shot setting. The best performances are highlighted in boldface.}
\label{ablation_qgr}
  \vspace{-10pt}
\end{table}

\subsection{Ablation Study}
\noindent{\textbf{Effect of Meta-Prompt Pre-training Module.}}
This module consists of three key components: meta-learning, text meta-prompts, and visual meta-prompts. We conduct ablation studies in Tab.~\ref{ablation_mera} to evaluate their zero-shot capabilities across various datasets. The baseline uses a CLIP model pre-trained on classification tasks. Row 2 evaluates meta-learning without prompts, where CLIP's visual and text encoders are fine-tuned to learn distortion knowledge across different distortions, improving IQA performance. However, tuning CLIP’s weights impairs its original generalization, resulting in lower performance on KonIQ. Rows 3 and 4 assess fine-tuning only text or visual prompts during meta-learning. These strategies notably boost zero-shot performance—particularly on real-world datasets like KonIQ and LIVEC—by preserving CLIP’s generalization while adapting to IQA tasks. Row 5 explores prompt tuning without meta-learning, underscoring meta-learning’s role in curbing overfitting and maintaining generalization. The strongest results emerge when all components are combined, highlighting their complementary strengths.

\begin{figure}[t]
  \centering
    \includegraphics[width=0.47\textwidth]{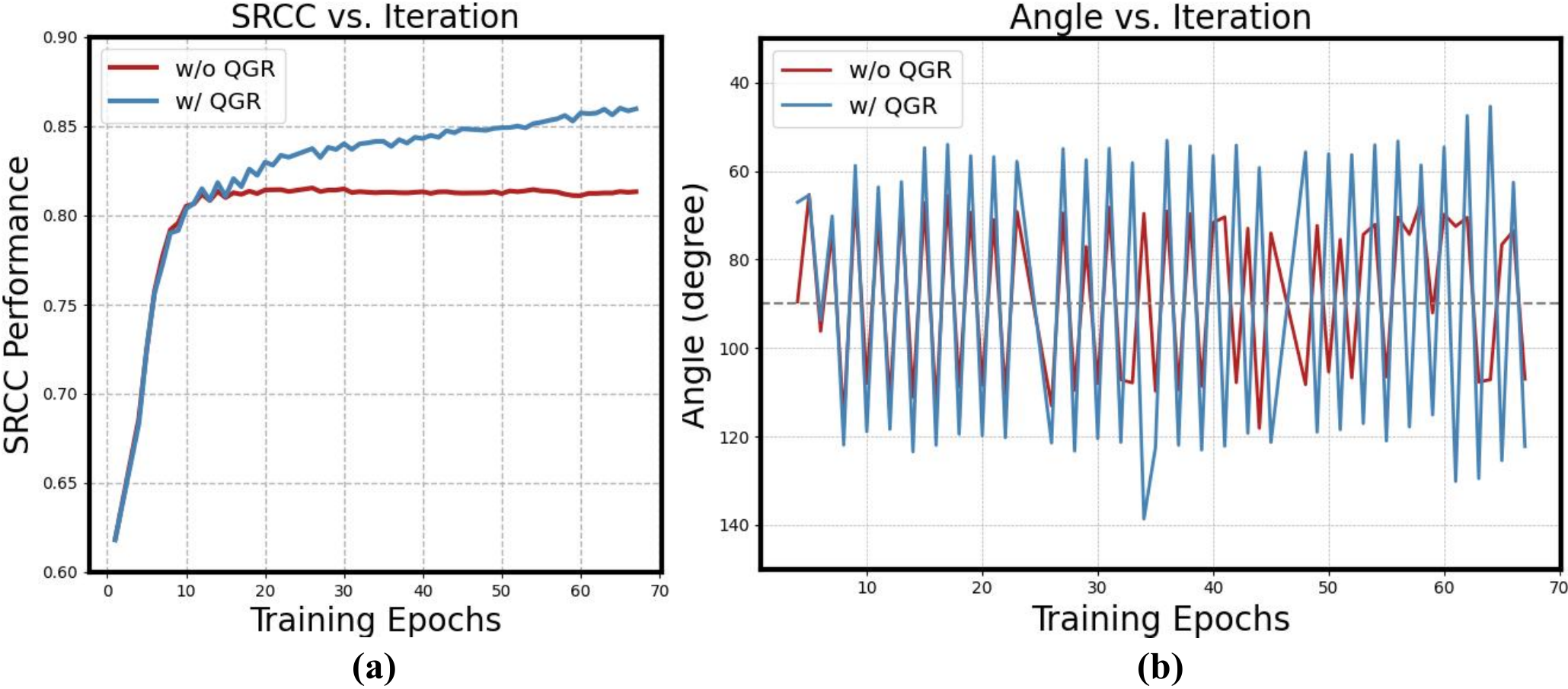}
     \vspace{-5pt}
  \caption{
(a) SRCC value during training with 50 samples. (b) Angles between $\boldsymbol{G}_{qua}$ and $\boldsymbol{G}_{sem}$ during training on KonIQ. Under the constraints of QGR, our method captures more correlation between quality knowledge and general semantic knowledge.}
  \label{qgr_angle}
       \vspace{-5pt}
\end{figure}

\begin{table}[t]
\centering
\resizebox{0.47\textwidth}{!}{
\begin{tabular}{ccccccc} 
\toprule
      & \multicolumn{2}{c}{50} & \multicolumn{2}{c}{100} & \multicolumn{2}{c}{200}  \\ 
\cmidrule{2-7}
$\lambda$ & PLCC  & SRCC           & PLCC  & SRCC            & PLCC  & SRCC             \\ 
\midrule
1     &0.850       &0.812      &0.874  &0.839           &0.893    & 0.866        \\
3     & 0.864 & 0.839          &0.890  &0.865  &0.903    &0.878        \\
\rowcolor{cyan!10} 5     & \textbf{0.880} & \textbf{0.853}          & \textbf{0.896} & \textbf{0.872}           & \textbf{0.908} & \textbf{0.883}   \\
7     & 0.876 & 0.851          &0.895 &0.870            &0.905 &0.881   \\
\bottomrule
\end{tabular}}
\caption{The ablation study about soft weight $\lambda$ in Eq.~\ref{eq_qgr}.}
\label{hyper}
  \vspace{-15pt}
\end{table}

\noindent{\textbf{Effect of Meta-learning.}} 
To further investigate whether the effectiveness of our method derives from meta-learning, we conducted an ablation study. Specifically, as shown in rows 1 and 2 of Tab.~\ref{ablation_meta}, when using Empirical Risk Minimization without meta-learning pre-training on two synthetic dataset, the fine-tuning performance on LIVEC and KonIQ dataset was even worse than the baseline without pre-training. This indicates that additional pre-training data does not necessarily enhance performance and may even lead to overfitting~\cite{li2024bridging}. In contrast, our meta-learning strategy effectively extracts generalizable quality priors from external data, significantly enhancing performance across datasets.

\noindent{\textbf{Effect of Gradient Regularization.}}
We conducted ablation studies on the QGR module. As shown in Tab.~\ref{ablation_qgr} and Fig.~\ref{qgr_angle}(a), without QGR, models are prone to overfitting under limited training data, with rapid performance saturation and a slight decline, which ultimately hinders generalization. By modulating training gradients, QGR effectively mitigates overfitting and enhances adaptability to the test dataset.
To illustrate QGR's impact, we analyzed the angular difference between gradients $\boldsymbol{G}_{\text{qua}}$ and $\boldsymbol{G}_{\text{sem}}$ during training, as shown in Fig.~\ref{qgr_angle}(b). Without QGR, the angle between $\boldsymbol{G}_{\text{qua}}$ and $\boldsymbol{G}_{\text{sem}}$ tends toward 90 degrees, reflecting the orthogonality typical of high-dimensional random vectors \cite{cai2013distributions}. In contrast, QGR introduces greater angular variation, suggesting that it encourages the model to explore more correlations between quality and semantic directions. This improves the model's ability to capture quality-related information and mitigates overfitting on limited data.

\noindent{\textbf{Effect of soft weight $\lambda$.}} On the KonIQ dataset, we conduct ablation experiments with various soft weights \(\lambda\) in Eq.~\ref{eq_qgr} to examine their impact, as detailed in Tab.~\ref{hyper}. Results show that a small \(\lambda\) diminishes the effectiveness of our QGR, whereas a large \(\lambda\) causes substantial gradient changes and decreases performance. Given our observation of the trade-off, we adopt $\lambda=5$ in our experiments.

\begin{figure}[t]
    \includegraphics[width=83mm, height=45mm]{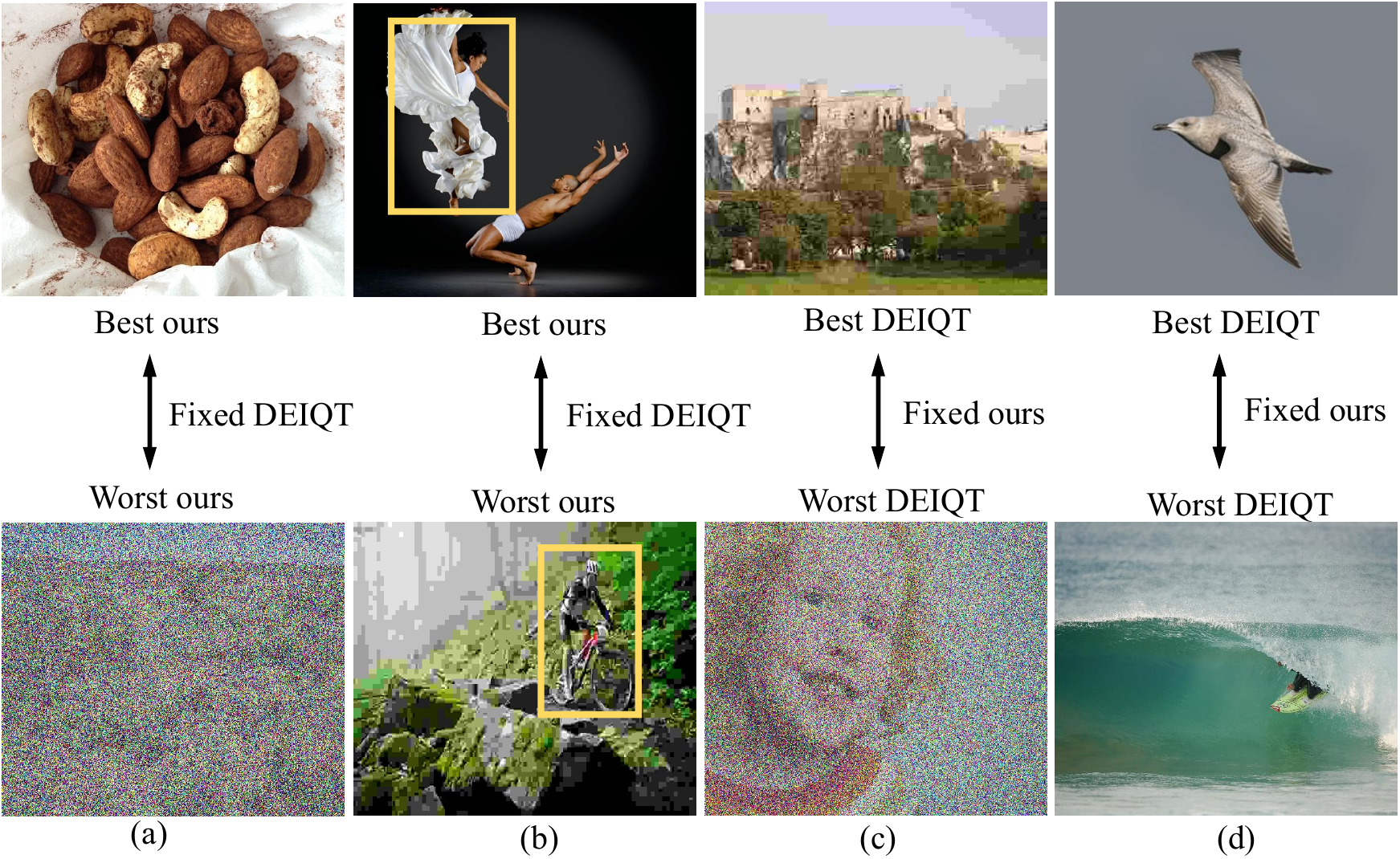}
     \vspace{-15pt}
	\caption{gMAD results between DEIQT~\cite{DEIQT} and our GRMP-IQA. (a) Fixed DEIQT at low quality. (b) Fixed DEIQT at high quality. (c) Fixed ours at low quality. (d) Fixed ours at high quality.
 }
 \label{gmad}
      \vspace{-5pt}
\end{figure}

\begin{table}[t!]
\centering
\renewcommand{\arraystretch}{1.15}
\setlength{\tabcolsep}{2.2pt} 
\resizebox{0.475\textwidth}{!}{
\begin{tabular}{c c c c cc}
\toprule
\multirow{2}{*}{{Method}} & \multirow{2}{*}{{Param}} & \multicolumn{2}{c}{{Efficiency}} & \multicolumn{2}{c}{{Performance}} \\
\cmidrule(lr){3-4} \cmidrule(lr){5-6}
& & {Throughput ↑} & {Latency ↓} & {80\% Train} & {20\% Train} \\
\midrule
Q-Align & 8.2B & 9.9 (img/sec) & 101.0 (ms) & 0.941 / \textbf{0.940} & 0.901 / 0.903 \\
\rowcolor{cyan!10}
Ours & \textbf{151M} & \textbf{43.0 (img/sec)} & \textbf{21.3 (ms)} & \textbf{0.945} / 0.934 & \textbf{0.931 / 0.915} \\
\bottomrule
\end{tabular}}
\caption{Comparison with Q-Align across varying training data ratio to evaluate efficiency and PLCC/SRCC on the KonIQ dataset.}
\label{result_pe}
 \vspace{-12pt}
\end{table}

\noindent{\textbf{Computational Analysis.}}
As shown in Tab.~\ref{result_pe}, we evaluate the inference latency and throughput of our method on an RTX3090 GPU, surpassing Q-Align~\cite{q-align} in both efficiency and accuracy. Notably, our model has only 41M learnable and 151M total parameters, far fewer than Q-Align's 8.2B.

\subsection{Qualitative Analysis}
The gMAD competition~\cite{ma2016group} is a standard method for assessing IQA model robustness. It selects image pairs where the attacker model predicts a large quality difference, while the defender perceives them as similar. Observers then evaluate these pairs to gauge model generalization.
As shown in Fig.~\ref{gmad}, when our model acts as the defender, the attacker’s selected pairs exhibit minimal perceptual quality changes. Conversely, as the attacker, our model consistently identifies pairs with significant quality differences. This highlights its strong defensive and offensive capabilities.
Notably, in the second column, DEIQT misclassifies semantically similar images as having comparable quality, whereas our model accurately distinguishes their differences.

\vspace{-3pt}
\section{Conclusion}
\vspace{-2pt}
In this paper, we propose the GRMP-IQA framework, which generalizes well with limited data. It includes a meta-learning pre-training module that enables the CLIP model to rapidly adapt to IQA tasks and an adaptive gradient regulation module that refines gradient trajectories during fine-tuning, focusing updates on quality-aware knowledge while minimizing the negative impact of over-reliance on semantic noise. Comprehensive experiments on various BIQA datasets validate the superior generalization of our framework, especially in data-scarce scenarios.

\noindent \textbf{Acknowledgements.}
This work was supported by the National Science Fund for Distinguished Young Scholars (No.62025603), the National Natural Science Foundation of China (No. U21B2037, No. U22B2051, No. U23A20383, No. U21A20472, No. 62176222, No. 62176223, No. 62176226, No. 62072386, No. 62072387, No. 62072389, No. 62002305 and No. 62272401), and the Natural Science Foundation of Fujian Province of China (No. 2021J06003, No.2022J06001).

{
    \small
    \bibliographystyle{ieeenat_fullname}
    \bibliography{main}
}

\end{document}